# Fine-Grained Word Sense Disambiguation Based on Parallel Corpora, Word Alignment, Word Clustering and Aligned Wordnets

**Dan TUFIŞ**
Institute for Artificial Intelligence
13, "13 Septembrie"
Bucharest, 050711
Romania
tufis@racai.ro

**Radu ION**
Institute for Artificial Intelligence
13, "13 Septembrie"
Bucharest, 050711
Romania
radu@racai.ro

**Nancy IDE**
Department of Computer Science, Vassar College
Poughkeepsie,
NY 12604-0520
USA
ide@cs.vassar.edu

**Abstract**

The paper presents a method for word sense disambiguation based on parallel corpora. The method exploits recent advances in word alignment and word clustering based on automatic extraction of translation equivalents and being supported by available aligned wordnets for the languages in the corpus. The wordnets are aligned to the Princeton Wordnet, according to the principles established by EuroWordNet. The evaluation of the WSD system, implementing the method described herein showed very encouraging results. The same system used in a validation mode, can be used to check and spot alignment errors in multilingually aligned wordnets as BalkaNet and EuroWordNet.

## 1   Introduction

Word Sense Disambiguation (WSD) is well-known as one of the more difficult problems in the field of natural language processing, as noted in (Gale et al, 1992; Kilgarriff, 1997; Ide and Véronis, 1998), and others. The difficulties stem from several sources, including the lack of means to formalize the properties of context that characterize the use of an ambiguous word in a given sense, lack of a standard (and possibly exhaustive) sense inventory, and the subjectivity of the human evaluation of such algorithms. To address the last problem, (Gale et al, 1992) argue for upper and lower bounds of precision when comparing automatically assigned sense labels with those assigned by human judges. The lower bound should not drop below the baseline usage of the algorithm (in which every word that is disambiguated is assigned the most frequent sense) whereas the upper bound should not be too restrictive" when the word in question is hard to disambiguate even for human judges (a measure of this difficulty is the computation of the agreement rates between human annotators).

Identification and formalization of the determining contextual parameters for a word used in a given sense is the focus of WSD work that treats texts in a monolingual setting—that is, a setting where translations of the texts in other languages either do not exist or are not considered. This focus is based on the assumption that for a given word $w$ and two of its contexts $C_1$ and $C_2$, if $C_1 \equiv C_2$ (are perfectly equivalent), then $w$ is used with the same sense in $C_1$ and $C_2$. A formalized definition of context for a given sense would then enable a WSD system to accurately assign sense labels to occurrences of $w$ in unseen texts. Attempts to characterize context for a given sense of a word have addressed a variety of factors:

- *Context length*: what is the size of the window of text that should be considered to determine context? Should it consist of only a few words, or include much larger portions of text?
- *Context content*: should all context words be considered, or only selected words (e.g., only words in a certain part of speech or a certain grammatical relations to the target word)? Should they be weighted based on distance from the target or treated as a "bag of words"?
- *Context formalization*: how can context information be represented to enable definitions of an inter-context equivalence function? Is there a single representation appropriate for all words, or does it vary according to, for example, the word's part of speech?

The use of multi-lingual parallel texts provides a very different approach to the problem of context identification and characterization.

"Context" now becomes the word(s) by which the target word (i.e., the word to be disambiguated) is translated in one or more other languages. The assumption here is that different senses of a word are likely to be lexicalized differently in different languages; therefore, the translation can be used to identify the correct sense of a word. Effectively, the translation captures the context as the translator conceived it.

The use of parallel translations for sense disambiguation brings up a different set of issues, primarily because the assumption that different senses are lexicalized differently in different languages is true only to an extent. For instance, it is well known that many ambiguities are preserved across languages (e.g. the French *intérêt* and the English *interest*), especially languages that are relatively closely related. This raises new questions: how many languages, and of which types (e.g., closely related languages, languages from different language families), provide adequate information for this purpose? How do we measure the degree to which different lexicalizations provide evidence for a distinct sense?

We have addressed these questions in experiments involving sense clustering based on translation equivalents extracted from parallel corpora ((Ide, 199; Ide et al., 2002). Tufiş and Ion (2003) build on this work and further describe a method to accomplish a "neutral" labeling for the sense clusters in Romanian and English that is not bound to any particular sense inventory. Our experiments confirm that the accuracy of word sense clustering based on translation equivalents is heavily dependent on the number and diversity of the languages in the parallel corpus and the language register of the parallel text. For example, using six source languages from three language families (Romance, Slavic and Finno-Ugric), sense clustering of English words was approximately 74% accurate; when fewer languages and/or languages from less diverse families are used accuracy drops dramatically. This drop is obviously a result of the decreased chances that two or more senses of an ambiguous word in one language will be lexicalized differently in another when fewer languages, and languages that are more closely related, are considered.

To enhance our results, we have explored the use of additional resources, in particular, the aligned wordnets in BalkaNet (Tufiş et al. 2004a). BalkaNet is a European project that is developing monolingual wordnets for five Balkan languages (Bulgarian, Greek, Romanian Serbian, and Turkish) and improving the Czech wordnet developed in the EuroWordNet project. The wordnets are aligned to the Princeton Wordnet (PWN2.0), taken as an interlingual index, following the principles established by the EuroWordNet consortium. The underlying hypothesis in this experiment exploits the common intuition that reciprocal translations in parallel texts should have the same (or closely related) interlingual meanings (in terms of BalkaNet, interlingual index (ILI) codes). However, this hypothesis is reasonable if the monolingual wordnets are reliable and correctly linked to the interlingual index (ILI). Quality assurance of the wordnets is a primary concern in the BalkaNet project, and to this end, the consortium developed several methods and tools for validation, described in various papers authored by BalkaNet consortium members (see Proceedings of the Global WordNet Conference, Brno, 2004).

We previously implemented a language-independent disambiguation program, called WSDtool, which has been extended to serve as a multilingual wordnet checker and specialized editor for error-correction. In (Tufiş, et al., 2004) it was demonstrated that the tool detected several interlingual alignment errors that had escaped human analysis. In this paper, we describe a disambiguation experiment that exploits the ILI information in the corrected wordnets

## 2   Methodology and the algorithm

Our methodology consists of the following basic steps:

1. given a bitext $T_{L1L2}$ in languages $L_1$ and $L_2$ for which there are aligned wordnets, extract all pairs of lexical items that are reciprocal translations: $\{<W^i_{L1}\ W^j_{L2}>^+\}$

2. for each lexical alignment $<W^i_{L1}\ W^j_{L2}>$, extract the ILI codes for the synsets that contain $W^i_{L1}$ and $W^j_{L2}$ respectively to yield two lists of ILI codes, $L^1_{ILI}(W^i_{L1})$ and $L^2_{ILI}(W^j_{L2})$

3. identify one ILI code common to the intersection $L^1_{ILI}(W^i_{L1}) \cap L^2_{ILI}(W^j_{L2})$ or a pair of ILI codes $ILI_1 \in L^1_{ILI}(W^i_{L1})$ and $ILI_2 \in L^2_{ILI}(W^j_{L2})$, so that $ILI_1$ and $ILI_2$ are the *most similar* ILI codes (defined below) among the candidate pairs $(L^1_{ILI}(W^i_{L1}) \otimes L^2_{ILI}(W^j_{L2}))$ [$\otimes$ = Cartesian product]

The accuracy of step 1 is essential for the success of the validation method. A recent shared task evaluation) of different word aligners (www.cs.unt.edu/~rada/wpt, organized on the occasion of the Conference of the NAACL showed that step 1 may be solved quite reliably. Our system (Tufiş et al. 2003) produced lexicons relevant for wordnets evaluation, with an aggregated F-measure as high as 84.26%. Meanwhile, the word-aligner was further improved so that current performance on the same data is about 1% better on all scores in word alignment and about 2% better in wordnet-relevant dictionaries. The word alignment problem includes cases of *null alignment,* where words in one part of the bitext are not translated in the other part; and cases of *expression alignment,* where multiple words in one part of the bitext are translated as one or more words in the other part. Word alignment algorithms typically do not take into account the part of speech (POS) of the words comprising a translation equivalence pair, since cross-POS translations are rather frequent. However, for the aligned wordnet-based word sense disambiguation we discard both translation pairs which do not preserve the POS and null alignments. Multiword expressions included in a wordnet are dealt with by the underlying tokenizer. Therefore, we consider only one-to-one, POS-preserving alignments.

Once the translation equivalents were extracted, then, for any translation equivalence pair $<W_{L1}\ W_{L2}>$ and two aligned wordnets, the steps 2 and 3 above should ideally identify one ILI concept lexicalized by $W_{L1}$ in language *L1* and by $W_{L2}$ in language *L2*. However, due to various reasons, the wordnets alignment might reveal not the same ILI concept, but two concepts which are semantically close enough to license the translation equivalence of $W_{L1}$ and $W_{L2}$. This can be easily generalized to more than two languages.

Our measure of interlingual concepts semantic similarity is based on PWN2.0 structure. We compute semantic-similarity score[1] by:

$$ss(ILI_1, ILI_2) = 1/1+k$$

where $k$ is the number of links from $ILI_1$ to $ILI_2$ or from both $ILI_1$ and $ILI_2$ to the nearest common ancestor. The semantic similarity score is 1 when the two concepts are identical, 0.33 for two sister concepts, and 0.5 for mother/daughter, whole/part, or concepts related by a single link. Based on empirical studies, we decided to set the significance threshold of the semantic similarity score to 0.33.

In order to describe the algorithm for WSD based on aligned wordnets let us assume we have a parallel corpus containing texts in $k+1$ languages ($T, L_1, L_2...L_k$), where $T$ is the target language and $L_1, L_2...L_k$ are the source languages and monolingual wordnets for each of the $k+1$ languages interlinked via an ILI-like structure. For each source language and for all occurrences of a specific word in the target language T, we build a matrix of translation equivalents as shown in Table 1 ($eq_{ij}$ represents the translation equivalent in the $i^{th}$ source language of the $j^{th}$ occurrence of the word in the target language).

|       | Occ #1     | Occ #2     | …   | Occ #n     |
|-------|------------|------------|-----|------------|
| $L_1$ | $eq_{11}$  | $eq_{12}$  | …   | $eq_{1n}$  |
| $L_2$ | $eq_{21}$  | $eq_{22}$  | …   | $eq_{2n}$  |
| …     | …          | …          | …   | …          |
| $L_k$ | $eq_{k1}$  | $eq_{k2}$  | …   | $eq_{kn}$  |

Table 1. The translation equivalents matrix (EQ matrix)

If the target word is not translated in language $L_i$, $eq_{ij}$ is represented by the null string.

The second step transforms the matrix in Table 1 to a VSA (Validation and Sense Assignment) matrix with the same dimensions (Table 2).

|       | Occ #1     | Occ #2     | …   | Occ #n     |
|-------|------------|------------|-----|------------|
| $L_1$ | $VSA_{11}$ | $VSA_{12}$ | …   | $VSA_{1n}$ |
| $L_2$ | $VSA_{21}$ | $VSA_{22}$ |     | $VSA_{22}$ |
| …     | …          | …          | …   | …          |
| $L_k$ | $VSA_{k1}$ | $VSA_{k2}$ | …   | $VSA_{kn}$ |

Table 2. The VSA matrix

Here, $VSA_{ij} = L^{EN}_{ILI}(W_{EN}) \cap L^i_{ILI}(W^j_{Li})$, where

---

[1] For other approaches to similarity measures see the discussion in Budanitsky and Hirst (2001)

$L^{EN}_{ILI}(W_{EN})$ represent the ILI codes of all synsets in which the target word $W_{EN}$ occurs, and $L^{i}_{ILI}(W^{j}_{Li})$ is the list of ILI-codes for all synsets in which the translation equivalent for the $j^{th}$ occurrence of $W_{EN}$ occurs.

If no translation equivalent is found in language $L_i$ for the $j^{th}$ occurrence of $W_{EN}$, VSA($i,j$) is undefined; otherwise, it is a set containing 0, 1, or more ILI codes. For undefined VSAs, the algorithm cannot determine the sense number for the corresponding occurrence of the target word. However, it is very unlikely that an entire column in Table 2 is undefined, i.e., that there is no translation equivalent for an occurrence of the target word in any of the source languages.

When VSA($i,j$) contains a single ILI code, the target occurrence and its translation equivalent are assigned the same sense.

When VSA($i,j$) is empty—i.e., when none of the senses of the target word corresponds to an ILI code to which a sense of the translation equivalent was linked--the algorithm selects the pair in $L^{EN}_{ILI}(W_{EN}) \otimes L^{i}_{ILI}(W^{j}_{Li})$ with the highest similarity score. If no pair in $L^{EN}_{ILI}(W_{EN}) \otimes L^{i}_{ILI}(W^{j}_{Li})$ has a the semantic similarity score above the significance threshold, neither the occurrence of the target word nor its translation equivalent can be semantically disambiguated; but once again, it is extremely rare that there is no translation equivalent for an occurrence of the target word in any of the source languages.

In case of ties, the pair corresponding to the most frequent sense of the target word in the current bitext pair is selected. If this heuristic in turn fails, the choice is made in favor of the pair corresponding to the lowest PWN2.0 sense number for the target word, since PWN senses are ordered by frequency.

When the VSA cell contains two or more ILI-codes, we have the case of *cross-lingual ambiguity*, i.e., two or more senses are common to the target word and the corresponding translation equivalent in the $i^{th}$ language. For example, at least two senses of the English word *movement* are identical to senses of the Romanian word *mişcare*. In these cases, the heuristics applied in the case of ties are applied.

### 2.1 Agglomerative clustering

As noted before, when VSA($i,j$) is undefined, we may get the information from a VSA corresponding to the same occurrence of the target word in a different language. However, this demands that aligned wordnets are available for all languages in the parallel corpus, and that the quality of the inter-lingual linking is high for all languages concerned. In cases where we cannot fulfill these requirements, we rely on a "back-off" method involving sense clustering based on translation equivalents, as discussed in (Ide, et al., 2002). We apply the clustering method after the wordnet-based method has been applied, and therefore each cluster containing an undisambiguated occurrence of the target word will also typically contain several occurrences that have already been assigned a sense. We can therefore assign the most frequent sense assignment in the cluster to previously unlabeled occurrences within the same cluster. The combined approach has two main advantages:

- it eliminates reliance only on high-quality, k-1 aligned wordnets. Indeed, having k+1 languages in our corpus, we need only apply the WSD method to the aligned wordnets for the target language (English in our case) and one source language, say $L_i$, and alignment lexicons from the target language to every other language in the corpus. The WSD procedure in the bilingual setting would ensure the sense assignment for most of the non-null translation equivalence pairs and the clustering algorithm would classify the target words which were not translated (or for which the word alignment algorithm didn't find a correct translation) in $L_i$ based on their equivalents in the other k-1 source languages.
- it can reinforce or modify the sense assignment decided by the tie heuristics in case of cross-lingual ambiguity.

To perform the clustering, we derive a set of *m* binary vectors $VECT(L_p, TW^i)$ for each source language $L_p$ and each target word *i* occurring *m* times in the corpus. To compute the vectors, we first construct a Dictionary Entry List $DEL(L_p,TW^i)=\{W^j \mid <TW^i, W^j>$ is a translation equivalence pair}, comprising the *ordered* list of all the translation equivalents in the source language $L_p$ of the target word $TW^i$. In this part of the experiment, the translation equivalents are automatically extracted from the parallel corpus using a hypothesis testing algorithm described in (Tufiş 2002). $VECT(L_p,TW^i_k)$ specifies which of the possible translations of $TW^i$ was actually used as an equivalent for the $k^{th}$ occurrence of $TW^i$. All positions in $VECT(L_p,TW^i_k)$ are set to 0 except the bit at position $h$, which is 1 if the translation equivalent $(L_p,TW^i_k) = DEL_h(L_p,TW^i)$. The vector for each target word occurrence is obtained by concatenating the $VECT(L_p,TW^i_k)$ for all $k$ souce languages and its length is

$$\sum_{p=1}^{k} |DEL(L_p,TW^i)|.$$

We use a Hierarchical Clustering Algorithm based on Stolcke's Cluster2.9 to classify similar vectors into sense classes. Stolcke's algorithm generates a clustering tree, the root of which corresponds to a baseline clustering (all the occurrences are clustered in one sense class) and the leaves are single element classes, corresponding to each occurrence vector of the target word. An *interior cut* in the clustering tree will produce a specific number (say X) of sub-trees, the roots of which stand for X classes each containing the vectors of their leaves. We call an interior cut a *pertinent cut* if X is equal to the number of senses $TW^i$ has been used throughout the entire corpus. One should note that in a clustering tree many pertinent cuts could be possible. The pertinent cut which corresponds to the correct sense clustering of the $m$ occurrences of $TW^i$ is called a *perfect cut*. However, if $TW^i$ has Y possible senses, it is possible that only a subset of the Y senses will be used in an arbitrary text. Therefore, a perfect cut in a clustering tree cannot be deterministically computed. Instead of deriving the clustering tree and guessing at a perfect cut, we stop the clustering algorithm when Z clusters have been created, where Z is the number of senses in which the occurrences of $TW^i$ have been used in the text in question. However, the value of Z is specific to each word and depends on the type and size of the text; it cannot therefore be computed *a priori*. In our previous work (Tufiş and Ion, 2003), to approximate Z we imposed an exit condition for the clustering algorithm based on distance heuristics. In particular, the algorithm stops when the minimal distance between the existing classes increases beyond a given threshold level:

$$\frac{dist(k+1) - dist(k)}{dist(k+1)} > \alpha \qquad (1)$$

where $dist(k)$ is the minimal distance between two clusters at the $k$-th iteration step and $\alpha$ is an empirical numerical threshold. Experimentation revealed that reasonable results are achieved with a value for $\alpha$ is 0.12. However, although the threshold is a parameter for the clustering algorithm irrespective of the target words, the number of classes the clustering algorithm generates (Z) is still dependent on the particular target word and the corpus in which it appears.

By using sense information produced by the ILI-similarity approach, the algorithm and its exit condition have been modified as described below:
- the sense label of a cluster is given by the majority sense of its members as assigned by the wordnet-based sense labeling; a cluster containing only non-disambiguated occurrences has an *any* sense label;
- two *joinable* clusters (that is the clusters with the minimal distance and the exit condition (1) not satisfied) are joint only when their sense labels is the same or one of them has an *any* sense label; in this case the *any* sense label is turned into the sense label of the sense-assigned cluster. Otherwise the next distant clusters are tried.
- the algorithm stops when no clusters can be joined anymore.

## 3   The Experiment

The parallel corpus we used for our experiments is based on Orwell's novel "Ninety Eighty Four" (1984) which has been initially developed by the Multext-East consortium. Besides Orwell's original text, the corpus contained professional translations in six languages (Bulgarian, Czech, Estonian, Hungarian, Romanian and Slovene). The Multext-East corpus (and other language resources) is maintained by Tomaž Erjavec and a new release of it may be found at http://nl.ijs.si/ME/V3. Later, the parallel corpus has been extended with many other new language translations. The BalkaNet consortium added three new translations to the "1984" corpus: Greek, Serbian and Turkish. Each language text is tokenized, tagged and sentence aligned to the English original. We extracted from the entire parallel corpus only the languages of concern in the BalkaNet project (English, Bulgarian, Czech, Greek, Romaniann, Serbian and Turkish) and further retained only the 1-1 sentence alignments between English and all the other languages[2].

The BalkaNet version of the "1984" corpus is encoded as a sequence of *translation units* (TU), each containing one sentences per language, so that they are reciprocal translations. In order to evaluate both the performance of the WSDtool and to assess the accuracy of the interlingual linking of the BalkaNet wordnets we selected a bag of English target words (nouns and verbs) occurring in the corpus. The selection considered only polysemous words (at least two senses per part of speech) implemented (and ILI linked) in all BalkaNet wordnets. There resulted 211 words with 1644 occurrences in the English part of the parallel corpus.

Three experts independently sense-tagged all the occurrences of the target words and the disagreements were negotiated until consensus was obtained. The commonly agreed annotation represented the Gold Standard (GS) against which the WSD algorithm was evaluated. Additionally, a number of 13 students, enrolled in a Computational Linguistics Master program, were asked to manually sense-tag overlapping subsets of the same word occurrences. The overlapping ensured that each target word occurrence was seen by at least three students. Based on the students' annotations, using a majority voting, we computed another set of comparison data which below is referred to as SMAJ (Students MAJority).

Finally, the same targeted words were automatically disambiguated by the WSDtool algorithm (ALG) which was run both with and without the back-off clustering algorithm. For the basic wordnet-based WSD we used the Princeton Wordnet, the Romanian wordnet and the English-Romanian translation equivalence dictionary. For the back-off clustering we extracted a four[3] language translation dictionary (EN-RO-CZ-BG) based on which we computed the initial clustering vectors for all occurrences of the target words.

Out of the 211 set of targeted words, with 1644 occurrences the system could not make a decision for 38 (18 %) words with 63 occurrences (3.83%). Most of these words were happax legomena (21) for which neither the wordnet-based step not the clustering back-off could do anything. Others, were not translated by the same part of speech, were wrongly translated by the human translator or not translated at all (28). Finally, four occurrences remained untagged due to the incompleteness of the Romanian synsets linked to the relevant concepts (that is the four translation equivalents had their relevant sense missing from the Romanian wordnet). Applying the simple heuristics (SH) that says that any unlabelled target occurrence receives its most frequent sense, 42 out of 63 of them got a correct sense-tag. The table below summarizes the results.

| WSD annotation | Precision | Recall | F |
| --- | --- | --- | --- |
| AWN | 74.88% | 72.01% | 73.41% |
| AWN + C | 75.26% | 72.38% | 73.79% |
| AWN + C + SH | 74.93% | 74.93% | 74.93% |
| SMAJ | 72.99% | 72.99% | 72.99% |

Table 4. WSD precision recall and F-measure for the algorithm based on aligned wordnets (AWN), for AWN with clustering (AWN+C) and for AWN+C and the simple heuristics (AWN+C+SH) and for the students' majority voting (SMAJ)

It is interesting to note that in this experiment the students' majority annotation is less accurate than the one achieved by the automatic WSD annotation in all three variants. This is a very encouraging result since it shows that the tedious hand-made WSD in building word-sense disambiguated corpora for supervised training can be avoided.

## 4 Conclusion

Considering the fine granularity of the PWN2.0 sense inventory, our disambiguation results using parallel resources are superior to the state of the art in *monolingual* WSD (with the same sense inventory). This is not surprising since the parallel texts contain implicit knowledge about the sense of an ambiguous word, which has been provided by human translators. The drawback of our approach is that it relies on the existence of parallel data, which in the vast majority of cases is not available. On the other hand, supervised monolingual WSD relies on the existence of large samples of training data, and our method can be applied to produce such data to bootstrap monolingual applications. Given that parallel resources are becoming increasingly available, in particular on the World Wide Web (see for instance http://www.balkantimes.com where the same news is published in 10 languages), and aligned wordnets are being produced for more and more languages, it should be possible to apply our and similar methods to large amounts of parallel data in the not-too-distant future.

---

[2] This way, we build a unique alignment for all the languages and, by exploiting the transitivity of sentence alignment, we are able to make experiments with any combination of languages.

[3] Although we used only RO, CZ and BG translation texts, nothing prevents us from using any other translations, irrespective of whether their languages belong or not to the BalkaNet consortium.

One of the greatest advantages of our approach is that it can be used to automatically sense-tag corpora in several languages at once. That is, if we have a parallel corpus in multiple languages (such as the Orwell corpus), disambiguation performed on any one of them propagates to the rest via the ILI linkage. Also, given that the vast majority of words in any given language are monosemous (e.g., approximately 82% of the words in PWN have only one sense), the use of parallel corpora in multiple languages for WSD offers the potential to significantly improve results and provide substantial sense-annotated corpora for training in a range of languages.

## Acknowledgements

The work reported here was carried within the European project BalkaNet, no. IST-2000 29388 and support from the Romanian Ministry of Education and Research.